\documentclass[]{spie}  

\usepackage{amsmath,amsfonts,amssymb}
\usepackage{array}
\usepackage{tabularx, booktabs}
\usepackage{graphicx}
\usepackage[colorlinks=true, allcolors=blue]{hyperref}

\title{Segment Any Medical Model Extended}

\author[ad]{Yihao Liu}
\author[ad]{Jiaming Zhang}
\author[b]{Andrés Diaz-Pinto}
\author[c]{Haowei Li}
\author[a]{Alejandro Martin-Gomez}
\author[de]{Amir Kheradmand}
\author[af]{Mehran Armand}
\affil[a]{Department of Computer Science, Johns Hopkins University, Baltimore, MD, USA}
\affil[b]{School of Biomedical Engineering \& Imaging Sciences, King’s College London, London, UK}
\affil[c]{Department of Biomedical Engineering, Tsinghua University, Beijing, China}
\affil[d]{Department of Neurology, Johns Hopkins University, Baltimore, MD, USA}
\affil[e]{Department of Neuroscience, Johns Hopkins University, Baltimore, MD, USA}
\affil[f]{Department of Orthopaedic Surgery, Johns Hopkins University, Baltimore, MD, USA}

\authorinfo{
Send correspondence to yliu333@jhu.edu
}

\begin{document} 
\maketitle

\begin{abstract}
The Segment Anything Model (SAM) has drawn significant attention from researchers who work on medical image segmentation because of its generalizability. However, researchers have found that SAM may have limited performance on medical images compared to state-of-the-art non-foundation models. Regardless, the community sees potential in extending, fine-tuning, modifying, and evaluating SAM for analysis of medical imaging.  An increasing number of works have been published focusing on the mentioned four directions, where variants of SAM are proposed. To this end, a unified platform helps push the boundary of the foundation model for medical images, facilitating the use, modification, and validation of SAM and its variants in medical image segmentation. In this work, we introduce SAMM Extended (SAMME), a platform that integrates new SAM variant models, adopts faster communication protocols, accommodates new interactive modes, and allows for fine-tuning of subcomponents of the models. These features can expand the potential of foundation models like SAM, and the results can be translated to applications such as image-guided therapy, mixed reality interaction, robotic navigation, and data augmentation.

\end{abstract}


\section{INTRODUCTION}
\label{sec:intro}  

Segmentation is a crucial task in computer vision. Image segmentation involves the division of an image into distinct regions, facilitating the analysis of sub-structures within the image. More specifically, semantic segmentation, assigning labels to each pixel of an image, is a key process in medical image analysis. It is widely used in the fields of radiology, pathology, endoscopy, and image-guided therapy. Example applications include tumor diagnosis \cite{ranjbarzadeh2021brain}, dental model reconstruction \cite{gan2017tooth}, jaw transplantation\cite{murphy2015optimizing}, brain mapping \cite{khan2008freesurfer}, lesion detection \cite{hashemi2018asymmetric}, bone abnormality identification \cite{shrivastava2020bone}, vascular structure analysis \cite{radaelli2010segmentation}, 2D/3D registration \cite{grupp2020automatic, otake2011intraoperative}, and fluoroscopy-guided procedures \cite{unberath2018deepdrr, gao2021fluoroscopic}.

Despite its wide use, the manual process of medical image segmentation can be laborious and tedious. This is mainly due to the high precision and quality requirement in clinical settings, the 3D nature of the volumetric medical image data, and the advanced knowledge required in understanding complex anatomy. Moreover, the variability and subjectivity between different annotators also pose challenges to the consistency and reliability of the tasks. To address these issues, automated segmentation techniques have been used, simplifying the process and reducing the cost in time and resources. The techniques employed have evolved over several decades. Early attempts used in segmentation are the basic thresholding and edge detection due to the limitation existed the computing power \cite{rosenfeld1976digital, canny1986computational}. More complex methods emerged later such as region growing, shape characteristics-based methods, active contours, and knowledge-based methods \cite{tang2000mri}. Along with those, water-shed algorithms \cite{ng2006medical}, morphological operations \cite{chudasama2015image}, and level set \cite{khadidos2017weighted} methods also gained attention. The recent decades marked a pivotal evolution of machine learning and deep learning techniques \cite{ng2006medical, pratondo2017integrating}. Among them, CNN, UNet \cite{li2018h}, and transformers \cite{cao2022swin, chen2021transunet} are widely used in medical image segmentation \cite{wang2022medical}.

Deep learning (DL) methods have exhibited promising potential, yet their generalization capabilities are limited. Models designed for medical image segmentation typically require training on specific datasets and are often tailored for a single task or anatomical region \cite{wang2020deep}. This limitation is a significant barrier to the practical application and clinical use of segmentation models. Training DL models requires specific hardware, and the medical image datasets used in training are typically sparse \cite{gao2023}. Recognizing these challenges, zero-shot methods and most recently, foundation vision models emerged as potential solutions, for their capability of evaluating test data of classes that have not been used in training.

Segment Anything Model (SAM) \cite{kirillov2023segment} is a notable foundation model among the large-scale models that have emerged for computer vision tasks. SAM presents a novel paradigm in the field of image segmentation for its strong capabilities for generating object masks under zero- or few-shot conditions \cite{xian2018zero}, where a model predicts masks based on images in the categories that were not observed/related during training . Our previous work \cite{liu2023samm}, Segment Any Medical Model (SAMM) has been proposed to integrate vanilla SAM models \footnote{In this paper, the term ``vanilla SAM'' is used to refer to the models released by Kirillov \textit{et al}\cite{kirillov2023segment}.} and 3D Slicer \cite{pieper20043d} for real-time, semi-automatic, modality-agnostic, and prompt-based medical image segmentation. It's been recognized that the performance of SAM on medical images \cite{zhang2023segment, mattjie2023exploring} needs further improvement as vanilla SAM may suffer from noticeable performance drops \cite{hu2023efficiently, shi2023cross, kellener2023utilizing, deng2023segment, roy2023sam, mazurowski2023segment, shi2023generalist, huang2023segment, cheng2023sam}. In the more recent literature, we has seen an increasing emergence of SAM variants and improvement techniques \cite{shi2023cross, zhang2023faster, bai2023sam++, chen2023sam, qiu2023learnable, shaharabany2023autosam}, using fine-tuning \cite{hu2023efficiently, bai2023sam++, qiu2023learnable, ma2023segment, paranjape2023adaptivesam, huang2023push, zhang2023customized}, validation \cite{mattjie2023exploring, deng2023sam}, augmentation \cite{zhang2023input}, and annotation methods \cite{huang2023push, lei2023medlsam, wang2023mathrm}. 

Open-source AI-assisted labeling framework has been proposed before \cite{diaz2022monai}, but not used in SAM and its variants. SAMME integrates MedSAM (first fine-tuned SAM) \cite{ma2023segment} and MobileSAM (most efficient) \cite{zhang2023faster}, and is adding more SAM variants and fine-tuning capabilities to the existing platform. Such fine-tuning capability within the platform of SAMME will enable convenient and interactive selections of models and their tuned components. 

Our current work, Segment Any Medical Model Extended (SAMME) aims to provide a unified platform based on SAMM \cite{liu2023samm}, with (1) integration of new SAM variants that have improved its performance, (2) improvements in real-time inference by using a more efficient and faster communication protocol, (3) extension of the interaction modes in all views, and (4) allowing for fine-tuning of components of the models. The repository for this work can be accessed in https://github.com/bingogome/samm.

\begin{figure}
    \centering
    \includegraphics[width=\textwidth]{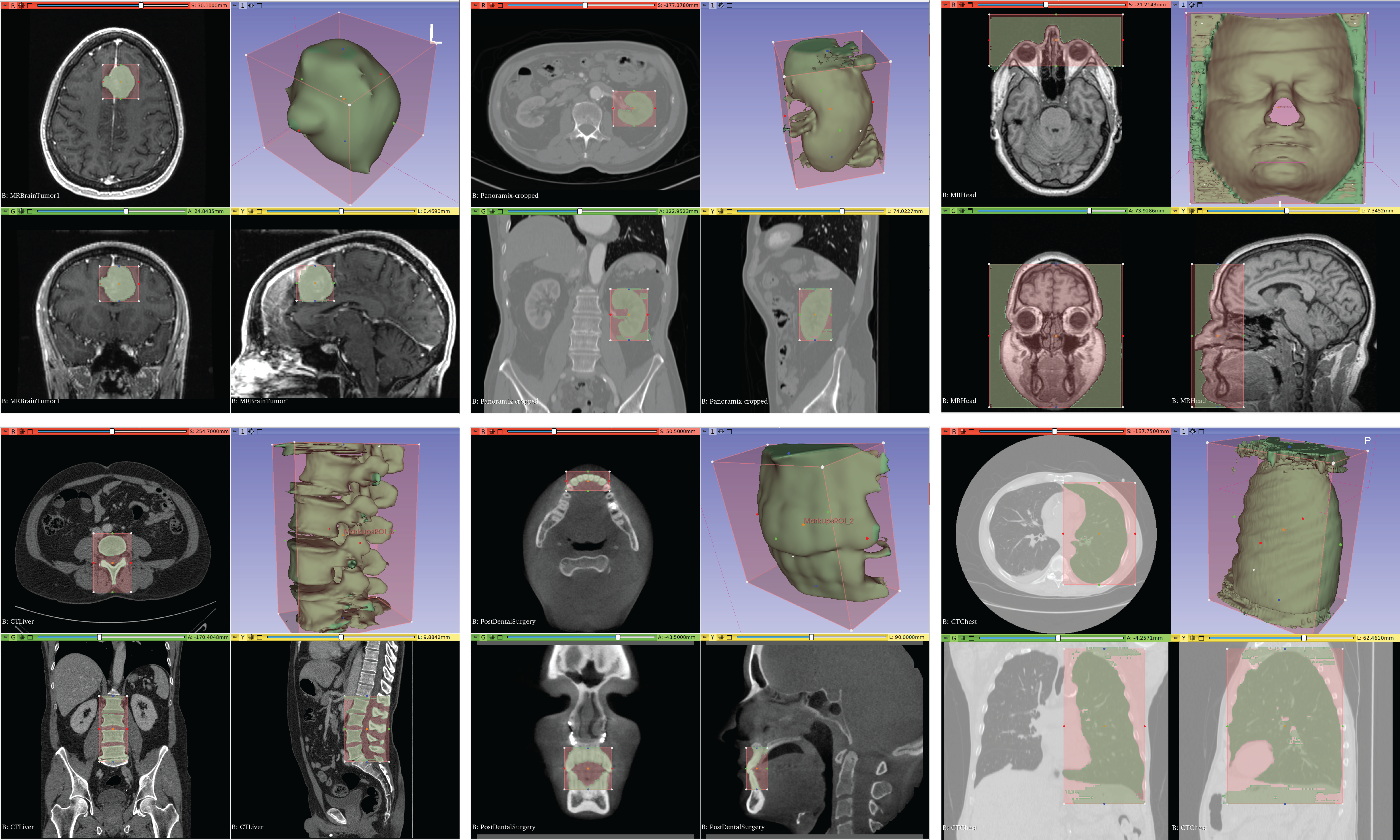}
    \caption{Examples of 3D bounding box prompts and the segmented 3D meshes using vanilla\_vit\_b model, which is a pretrain model provided along with original SAM. Datasets are from 3D Slicer sample datasets, listed sequentially, reading from left to right: MRBrainTumor1, CTA Abdomen, MRHead, CTLiver (here used for spine segmentation), CBCTDentalSurgery, and CTChest.}
    \label{fig:3dbbox}
\end{figure}

\begin{figure}
    \centering
    \includegraphics[width=0.8\textwidth]{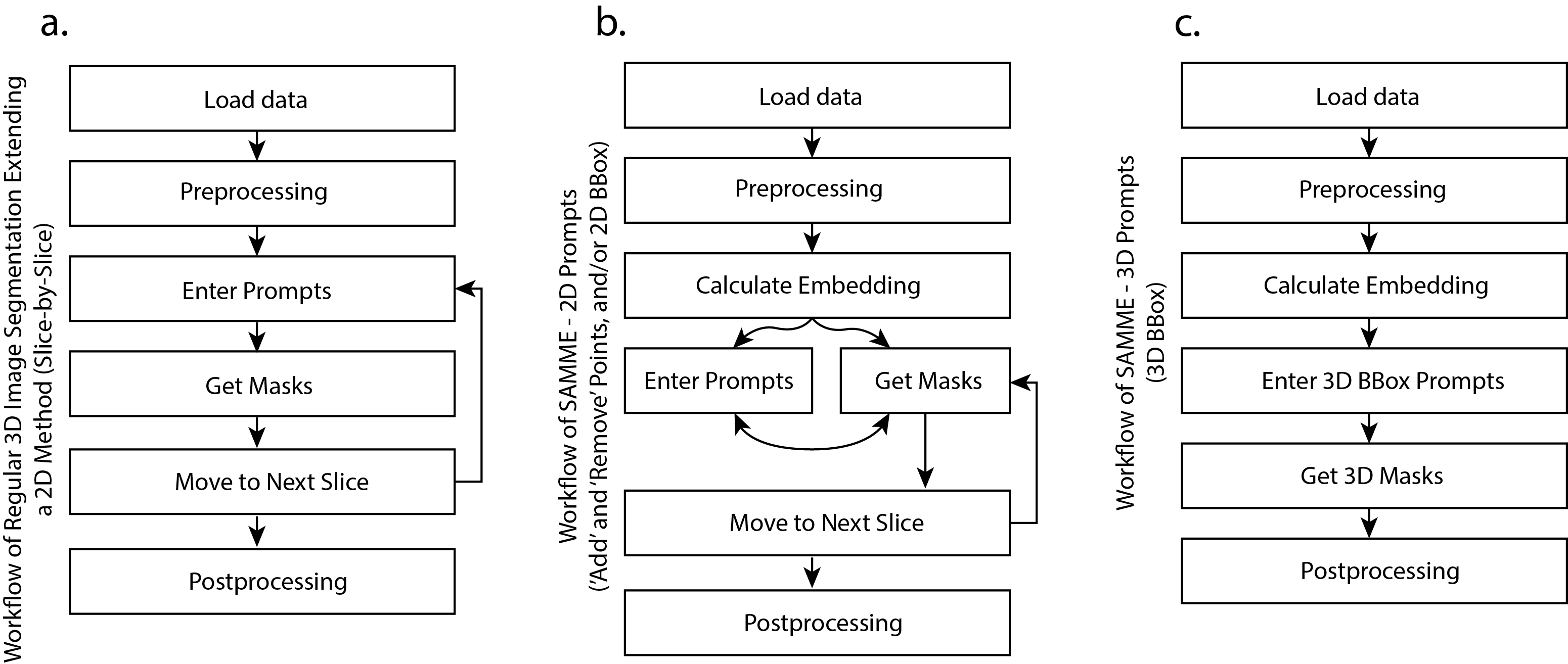}
    \caption{(a). The workflow for regular volumetric segmentation using 2D image segmentation tools. (b). The workflow for segmenting in SAMME using 2D prompts. (c). The workflow for segmenting in SAMME using 3D bounding boxes. In SAMME, the mask inference using 2D prompts is in realtime. At each cycle, the prompts are synchronized with the mask inference, so the ``Enter Prompts'' and ``Get Masks'' are iterative. The mask inference using 3D prompts is automated, so the workflow only takes in prompts once. }
    \label{fig:workflow}
\end{figure}

\begin{figure}
    \centering
    \includegraphics[width=\textwidth]{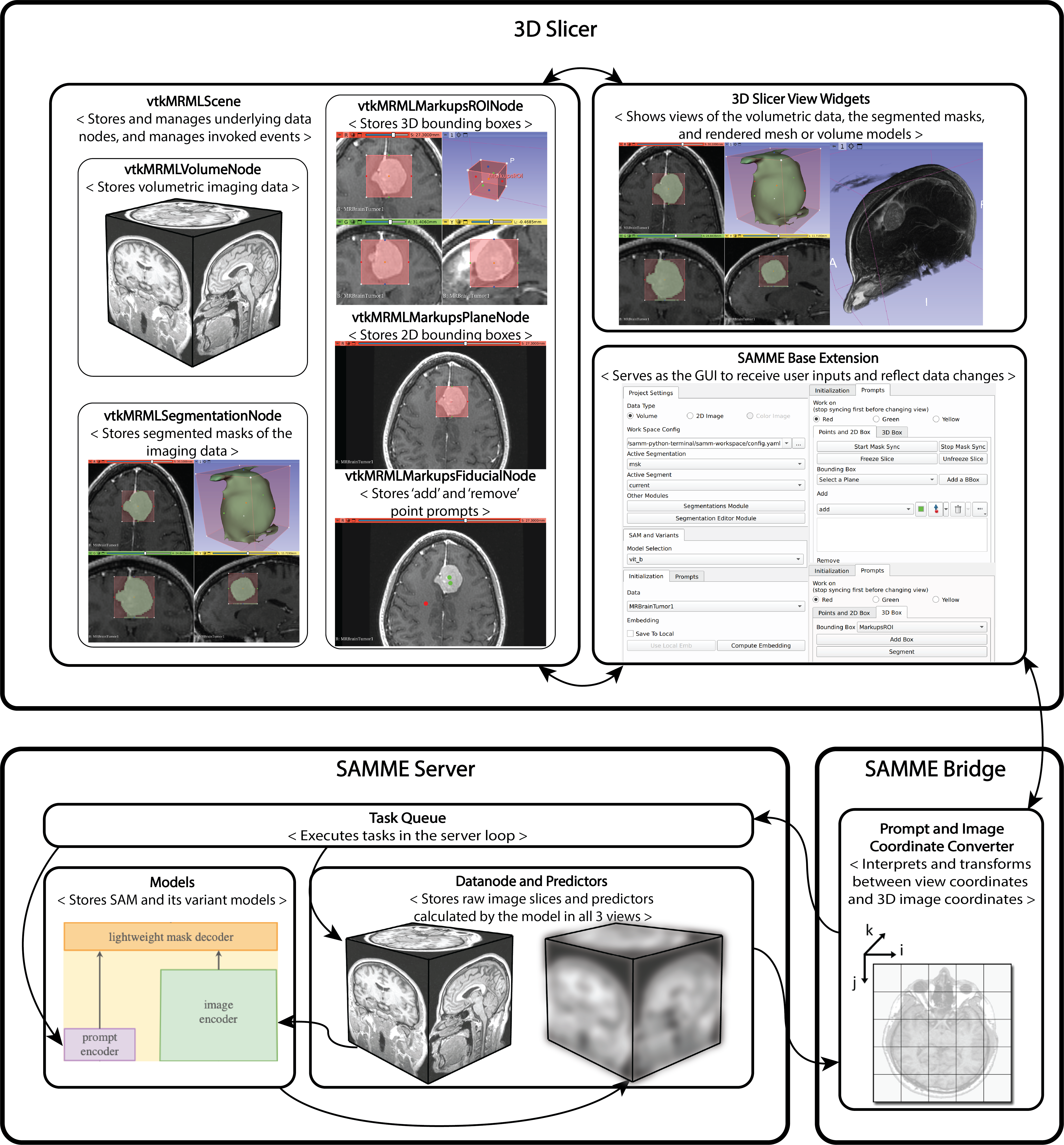}
    \caption{The architecture of SAMME. The 3D Slicer components handle the data storage, visualization, user interaction, as well as additional off-the-shelf functionalities. The SAMME Server runs the task queue which performs model computations and mask predictions. The SAMME Bridge interprets and converts the image coordinates data. The purposes of subcomponents are indicated in the corresponding angle brackets. The arrows in the figure are the communication channels between components.}
    \label{fig:architecture}
\end{figure}

\section{Methods}

The use of SAM in medical images presents two challenges inherent to SAM's nature. First, SAM has a scope for 2D images and is not intended for 3D image segmentation. The primary use of SAM in medical imaging is to efficiently extend a 2D tool for segmenting 3D data, given that most medical image modalities such as MRI and CT, are predominantly in volumetric format. Second, SAM aims for semi-automatic segmentation, relying on manual prompts \cite{kirillov2023segment}. Additionally, even though SAM provides fully automatic segmentations of 2D images, the proposed segmented masks by the model rely on additional selection logic for the targeted tasks to determine the final segmentation. The main reason is that the model is not trained on the target dataset or segmentation tasks. The semi-automatic nature of SAM poses challenges for a slice-by-slice approach, as the annotator needs to enter prompts manually for each slice, which is laborious and tedious.  Even though SAMME also adopts a slice-by-slice approach, the focus here is on traversing an anatomical axis efficiently.

This section describes the design of SAMME to address these problems: (1) By providing real-time segmentation capability, SAMME supports fast inference. (2) It also allows prompt propagation to eliminate the need for manual entry of the prompts for each slice. (3) SAMME supports the creation of 3D bounding boxes, requiring the annotators to enter only one 3D prompt for each target sub-structure in a volumetric image (Figure \ref{fig:3dbbox}). Aside from these solutions, the extension also includes the integration of recent SAM variants and their fine-tuning capabilities including adding medical images to the training dataset and training the mask decoder for medical image purposes.

\subsection{Segmentation Process}

Here we refer ``regular segmentation processes'' to be the methods that extend a 2D method for 3D segmentation, without exploiting the similarity between close slices. Figure \ref{fig:workflow} illustrates the segmentation process of volumetric images using both the regular method and SAMME. While both methods employ a slice-by-slice approach, there are notable differences between them. 

In a regular segmentation process, the typical procedure involves annotating a slice and moving to the next. This, however, is not efficient in an interactive method using SAM as the backbone model. As a foundation model, SAM has a large size so the processing time can be significant. To address this, SAMME uses a step to precompute the embedding of all slices in a volumetric image data. This precomputation process of the embedding is also performed in all 3 axes so it can support the prompting in all 3 anatomical views. The implementation of the process is explained in Section \ref{sec:arc} and the architecture is shown in Figure \ref{fig:architecture}.

Another difference is in the timing of obtaining the masks. In regular methods, masks are generated after the prompts are entered, while SAMME synchronizes the steps of entering prompts and predicting masks. The real-time capability of SAMME results in minimal waiting times (less than 0.1 seconds, shown in the section \ref{sec:result}), ensuring almost instantaneous responses to prompts. 

The manual entry of prompts may not be necessary when segmenting 3D image slice by slice, especially when consecutive slices are similar. Examples are shown in Figure \ref{fig:propandwl}a., where close slices tend to be alike, indicating that prompts for segmenting such slices do not require significant changes. Prompt propagation is introduced in SAMME to address this issue, which features the efficient reuse of the same prompts for multiple adjacent slices along an anatomical view.

\subsubsection{Prompt Propagation}\label{sec:prop}

The prompt propagation is the process of reusing the same input along an anatomical axis when segmenting similar slices. The implementation of prompt propagation when using 2D prompts, including ``add'', ``remove'' points and bounding boxes, improves the efficiency of segmentation. In SAMME, the user can simply scroll the mouse to move to the next slice, while prompts being propagated and the mask generation being synchronized. As prompts are propagated, the mask of the next slice is automatically updated in the visualization widgets of medical imaging software. Examples are shown in Figure \ref{fig:propandwl}a., where coronal, axial, and sagittal views of the segmentation of a brain tumor are presented. Along each axis, 5 slices of the corresponding view are displayed in a range of 20 mm. Each slice of these brain tumor images are similar, so the prompts can be reused. In a medical image volume with a voxel size of $1mm \times 1mm \times 1mm$, a distance of 20 mm in one axis corresponds to 20 slices. This means that in a semi-automated annotation workflow, the annotator needs to label the pixels in each of these 20 slices. However, with prompt propagation and real-time inference in SAMME, labeling a substructure, as demonstrated in Figure \ref{fig:propandwl}a., only takes few seconds.

In the workflow in Figure \ref{fig:workflow}b, prompt propagation corresponds to the cycle between ``Get Masks'' $\rightarrow$ ``Move to Next Slice'', because prompts are reused. While in regular methods (Figure \ref{fig:workflow}a), the cycle needs to be ``Enter Prompts'' $\rightarrow$ ``Get Masks'' $\rightarrow$ ``Move to Next Slice''. This simplification eliminates the redundant prompting step for slices are are close to the first slice where the prompts are initially entered.

\subsubsection{3D Bounding Box}

SAMME also supports segmeting volume using 3D bounding boxes, which is a fully automatic process except for the initial 3D bounding box input. Annotators can draw bounding boxes in the 3 anatomical views of the graphical user interface in 3D Slicer. Once defined, the segmentation of all slices within the 3D bounding box is automatically predicted. Since the process of the segmentation of a volume using 3D bounding boxes is essentially a propagated 2D bounding box along an axis, users can choose the view along which the automatic segmentation is performed. The size of the bounding box can be adjusted by moving its boundaries. 

Figure \ref{fig:3dbbox} illustrates examples of segmentation using a 3D bounding box in 3 views. The segmented mesh can be directly visualized on the rendering widget of 3D Slicer and can be exported for further evaluation as needed in a medical workflow. The examples in Figure \ref{fig:3dbbox} include various anatomies such as brain tumors, kidneys, facial skin (negative), spine, teeth, and lungs. Each edge and corner of the 3D bounding box can be interactively adjusted and the segmentation updates dynamically as adjustments are made, allowing for efficient adjustments to the segmentation.

\subsection{Architecture}\label{sec:arc}

The SAMME architecture closely follows the pattern established by SAMM \cite{liu2023samm}, as depicted in Figure \ref{fig:architecture}. SAMME is composed of three major components: 3D Slicer, SAMME Server, and SAMME Bridge.  Each major component has subcomponents serving for various purposes. 3D Slicer is an open-source software known for extendability of its visualization and image processing functionalities for other applications \cite{pieper20043d}. Widely used in medical image processing, robotics, and mixed reality research \cite{liu2023toward, li20233d, li2023evd}, 3D Slicer offers off-the-shelf functionalities \cite{fedorov20123d, pinter2012slicerrt, kikinis20133d}. SAMME uses the view widgets to visualize medical image views, segmented masks, and render reconstructed meshes or volumes. The underlying data is managed by Visualization Toolkit (VTK) packages \cite{schroeder2000visualizing} and scenes in Medical Reality Modeling Language (MRML) \cite{fedorov20123d}. The \textit{SAMME Base Extension} facilitates the integration of MRML scene, VTK functionalities, and 3D Slicer's existing features for rendering, visualization, and data processing. All user configurations, including the selection of volume images, segmentation nodes, SAM variant models, the view to work on, and input prompts, are entered in the graphical user interface within the extension.

The primary objective of the architecture is to achieve efficient segmentation. Prompt propagation (Section \ref{sec:prop}) simplifies the segmentation process, which requires fast inference of a single slice of a volume. Due to the size of the SAM model, computing the mask of a slice end-to-end upon request is not feasible. Therefore, SAMM \cite{liu2023samm} pre-computes the embeddings (features) of all slices in a volume and stores them in memory. Such save-and-retrieve paradigm ensures efficiency during inference. SAMME inherits this design and isolates the \textit{Models} component (Figure \ref{fig:architecture}) in the \textit{SAMME Server}, allowing for easy integration of any model following the same ``prompt encoder - image encoder - mask decoder'' architecture. SAM's API includes a ``SamPredictor'' \footnote{Repository: https://github.com/facebookresearch/segment-anything/blob/main/segment\_anything/predictor.py} class that encapsulates the model components, making function calls straightforward, which SAMME exploits to compute and store the embeddings efficiently. All tasks that are SAM model related are pipelined in the \textit{Task Queue} subcomponent in the \textit{SAMME Server}.

\subsection{Integration of the SAM Variants}

SAMME includes not only the vanilla SAM but also recent fine-tuned models. These recent models include lightweight or fine-tuned subcomponents. Among the models, notably, MobileSAM aims for a light-weight and efficient structure, which reduces the size of the image encoder of the vanilla SAM, by employing decoupled distillation between the optimization of the vanilla image encoder and mask decoder \cite{zhang2023faster}. MedSAM aims to fine-tune vanilla SAM for better segmentation on medical image domain \cite{ma2023segment}. 

\begin{figure}
    \centering
    \includegraphics[width=\textwidth]{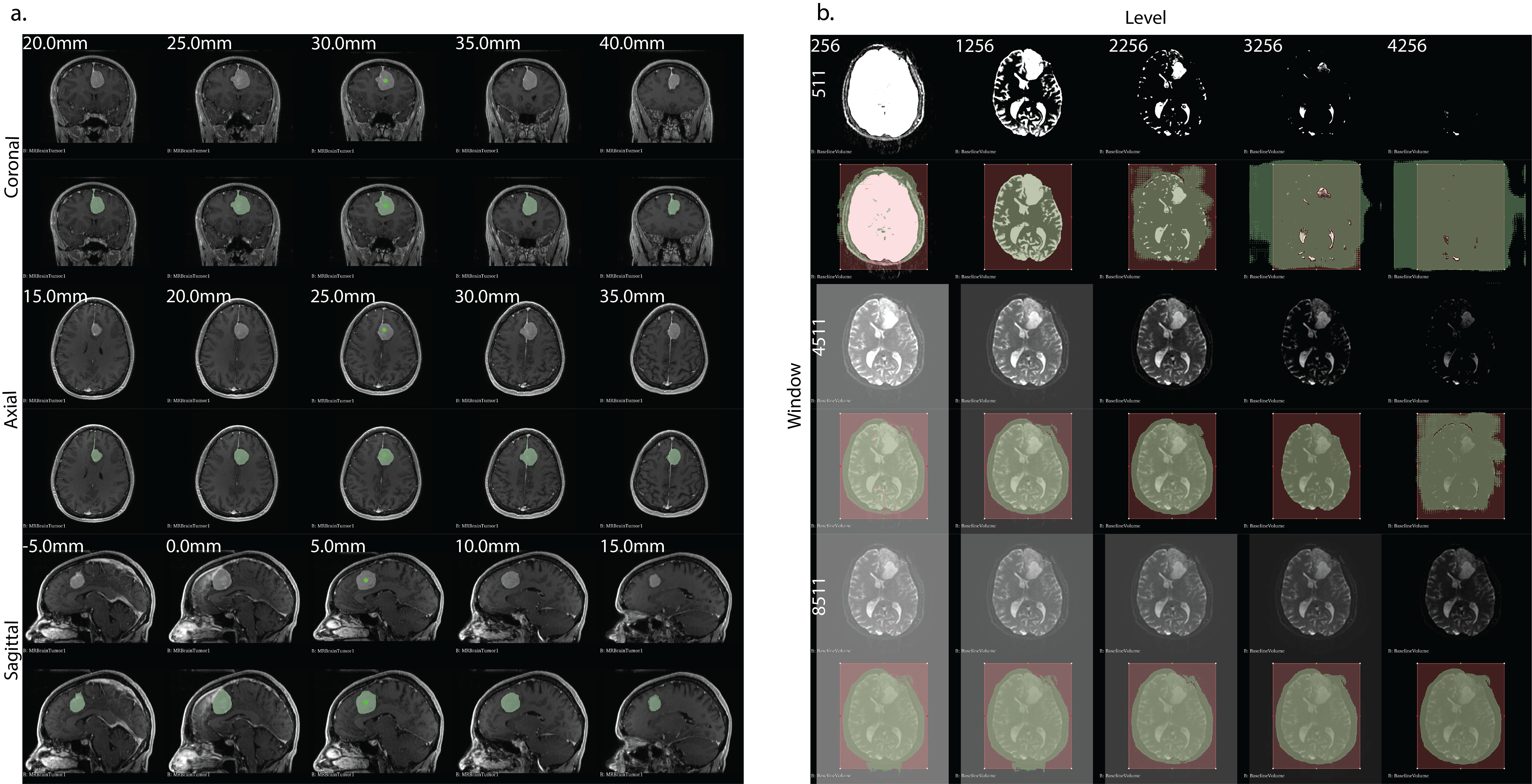}
    \caption{(a) Prompt propagation on close slices along 3 views of medical images (3D Slicer sample data MRBrainTumor1). Segmentation of a volumetric image can be significantly simplified because of the similarity between close slices. The positions of the slices are shown at the top left corner of each raw image, ranging 20 mm in each view. This figure shows that the same prompt along a view can be propagated, and close slices do not need the re-entry of the prompts. In SAMME, this is done simply by scrolling the mouse, and the prompt will be synchronized to the next slice. (b) Inference results using the same 2D bounding box with different window and level values (3D Slicer sample data BaselineVolume). Different inference results demonstrate the performance discrepancies, indicating window and level values should be considered during training or fine-tuning of SAM variants. All tests in this figure use the vanilla\_vit\_b model.}
    \label{fig:propandwl}
\end{figure}

The integration of fine-tuned models supports the goal of SAMME to explore and adopt a broader range of models for medical image analysis. To integrate a new SAM variant into SAMME, the use of a ``SamPredictor'' \cite{kirillov2023segment} class is required. This class encapsulates the SAM and its variants' model structure and provides access to the model weights. SAMME uses the predictor class to ensure that the precomputation step for embedding can occur without modifying the new SAM variant. While there may be minor differences in the preset of the models such as image size or pixel intensity, these can be addressed in the preprocessing. If required by a model, the preprocessing can be integrated into the image encoder. For instance, in MedSAM \cite{ma2023segment}, all image regardless of modality are resized to 1024 by 1024, and pixel intensity changes are applied. In the SAMME implementation, the original size and intensity of the medical image are maintained, but the interfaces to adjust window and level values are provided. A set of different values can potentially lead to different segmentation results (Figure \ref{fig:propandwl}b). In summary, any SAM variant can be integrated to SAMME if the following conditions are met:
\begin{itemize}
    \item The trained weights are provided.
    \item The inference can be performed using the ``SamPredictor'' class interface provided by SAM.
    \item The general architecture adheres to the ``prompt encoder - image encoder - mask decoder'' paradigm.
\end{itemize}

\begin{figure}
    \centering
    \includegraphics[width=0.6\textwidth]{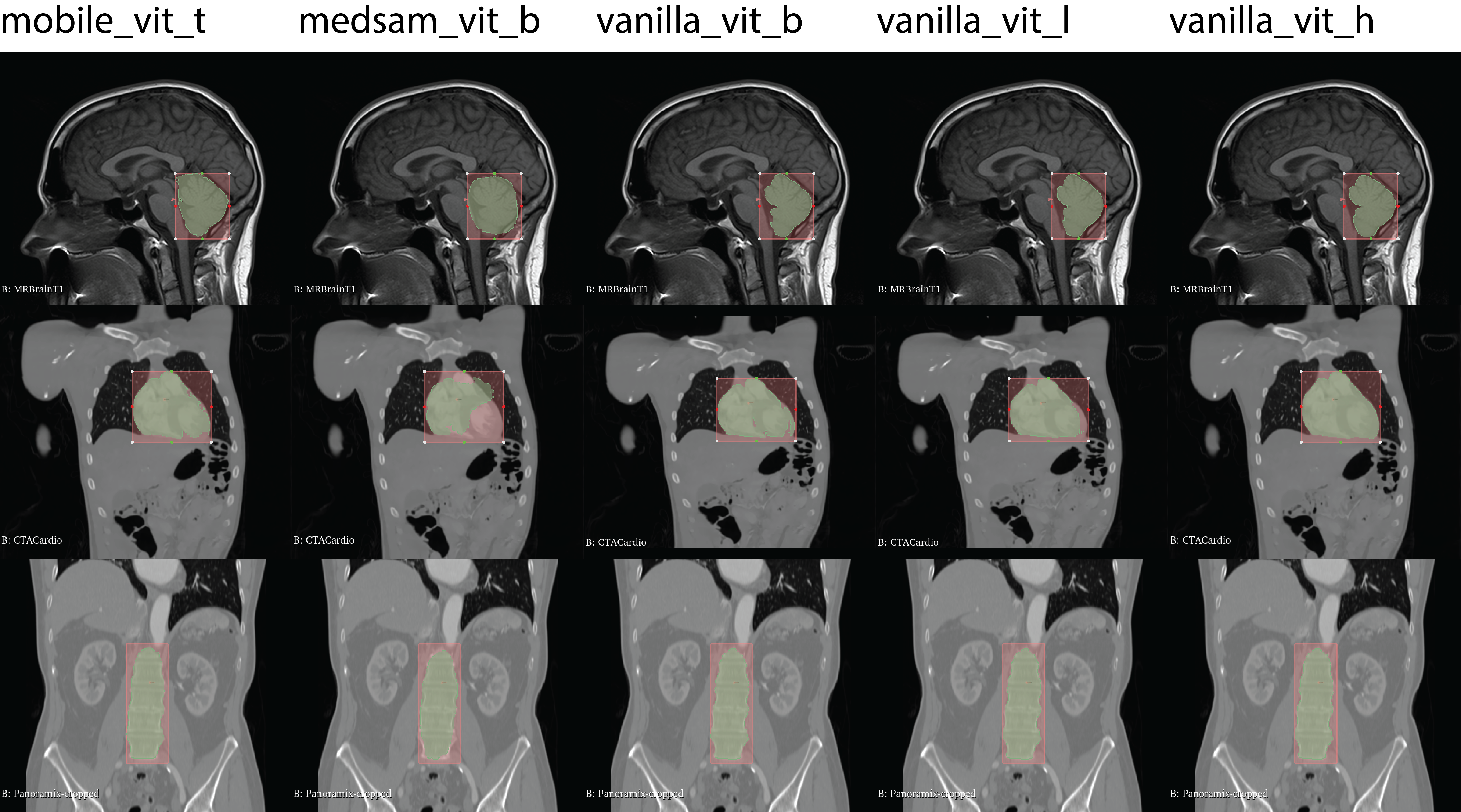}
    \caption{Segmentation results of 5 different SAM models using the same 2D bounding box. The dataset is from 3D Slicer sample data: CT-MR Brain, CTACardio, and CTA Abdomen (Panoramix).}
    \label{fig:modelseg}
\end{figure}

\section{Results}\label{sec:result}

Inherited from SAMM \cite{liu2023samm}, SAMME achieves real-time segmentation by using a save-and-retrieve paradigm for the embeddings of image slices. SAMME adopts a similar but improved architecture for faster real-time mask inference, that can be used with both the vanilla SAM and variants models. The improved architecture allows real-time mask inference of 0.008 seconds, with a complete cycle time of 0.06 seconds, 10 times as fast as previously implemented SAMM. And the save-and-retrieve paradigm is modified to pre-compute embeddings on three views on 3D Slicer, allowing flexible real-time interactive segmentation when there is different anatomy accessibility between views.

The real-time capability is enhanced by an improved and stable communication method between the segmentation server and 3D Slicer, allowing (1) direct retrieval of embeddings from the memory, and (2) efficient transmission of computer masks. On a test volume (256×256×130) and a Ubuntu 20.04 machine (AMD Ryzen 9 3900X, Nvidia GeForce RTX 3090), SAMM uses 0.6 seconds for a complete cycle from sending the inference request to overlaying the inferred mask to the medical image \cite{liu2023samm}, while SAMME has a time cycle of about 0.06 seconds across all the five tested models (Table 1). These models include pre-trained MobileSAM \cite{zhang2023faster} and MedSAM \cite{ma2023segment}, the lightest SAM and the first fine-tuned SAM, respectively.  SAMME also supports accessible interfaces for training or fine-tuning models on any user-specified datasets.

\begin{table}
\caption{Efficiency of inference in realtime (unit: seconds)}
\label{tab:efficiency}
\newcolumntype{Y}{>{\centering\arraybackslash}X}
\begin{tabularx}{\textwidth}{lYYYYY}
\toprule
\textbf{Models}& mobile\_vit\_t& medsam\_vit\_b& vanilla\_vit\_b& vanilla\_vit\_l& vanilla\_vit\_h\\
\midrule
\textbf{Avg. mask overlay time}&0.052 &0.051 &0.050 &0.050 &0.051\\
\textbf{Avg. inference time}&0.008 &0.008 &0.008 &0.008 &0.007\\
\textbf{Embedding calculation time}&7.231 &35.33 &35.436 &85.246 &146.783\\
\bottomrule
\end{tabularx}
\end{table}

SAMME presents an integration of SAM and its variants to support easier integration into the medical image processing. The segmentation results of sample data are presented in Figure \ref{fig:3dbbox} (3D bounding boxes), Figure \ref{fig:propandwl} (segmenting on 3 views and effects of volume window and level values), and Figure \ref{fig:modelseg} (different models). The models, including the vanilla models and the SAM variants, are non- to minimally-modified. Thus, the segmentation results are the same when using the equivalent configurations as indicated in the released models. The numerical accuracies of these models can be referenced, although the configuration of the experiments should be noted.

SAMME extends the interactive mode of SAMM \cite{liu2023samm} from only using ``add'' and ``remove'' point prompts to 2D bounding boxes or a combination of both. Selection with a 3D bounding box and a fully automatic generation of 3D masks without prompting is also possible. Instead of using single-view interactive segmentation, SAMME can perform segmentation on all three views on 3D Slicer. Figure \ref{fig:3dbbox} and Figure \ref{fig:propandwl}a show the segmentations using 3D bounding boxes and 2D bounding boxes. The propagation in 3 different anatomical views is also illustrated.  During the real-time inference, the prompts can be propagated along the view axis, achieving semi-automatic annotation. Using existing features of 3D Slicer, the results could be translated to other applications, such as image-guided therapy, mixed reality interaction, robotic navigation, and data augmentation.

Different models are used to test the integration with SAMME. Figure \ref{fig:modelseg} shows the example segmentation results using mobile\_vit\_t \cite{zhang2023faster}, medsam\_vit\_b \cite{ma2023segment}, vanilla\_vit\_b, vanilla\_vit\_l, and vanilla\_vit\_h \cite{kirillov2023segment}. Segmentation differences can be observed from the figure, which may be caused by the preprocessing differences in the images, as illustrated in examples in Figure \ref{fig:propandwl}b. In particular, the results presented in Figure \ref{fig:modelseg} are all using the window and level values automatically preset in 3D Slicer. In addition, SAMME does not resize the images, while the SAM variants from the community may have resizing steps in the preprocessing of the image slices. For example, all images are resized to 1024 $\times$ 1024 in MedSAM \cite{ma2023segment}. 

The prompting methods (bounding boxes and prompt points) are subjective, and the literature has not investigated a consistent way of prompting. Aside from prompting, volume window and level values are also subjective, which may also cause different segmentation results. In practice, setting the window and level values are done in preprocessing. We demonstrate the effects of volume window and level values in Figure \ref{fig:propandwl}b. 3D Slicer provides off-the-shelf adjustment interfaces, so the preprocessing step is integrated seemlessly.

\section{CONCLUSION}

The rise of vision foundation models presents an opportunity for more efficient semi-automated medical image segmentation. SAM is the most noteworthy foundation model featuring zero- or few-shot capability. The generalizability of this foundation model is promising, eliminating the need for extended training. The objective of our work is to facilitate the application of state-of-the-art models in the medical image segmentation process. This work extends the previous routines of coordinate manipulation, prompt annotation, modality-agnostic segmentation, and image-mask overlay visualization. The overall architecture that consists of a SAMME server and an interactive plugin for 3D Slicer was optimized to achieve real-time mask generation. The integration of the most recent variants of SAM will allow a comparison of new methods, not limited to annotation, fine-tuning, and validation. 

\section{Acknowledgments}
This work was supported by grants from the National Institute of Deafness and Other Communication Disorders (R01DC018815), National Institute of Arthritis and Musculoskeletal and Skin Diseases (R01AR080315), and National Institute of Biomedical Imaging and Bioengineering (R01EB023939).

\bibliography{report} 
\bibliographystyle{spiebib} 

\end{document}